\documentclass[10pt, a4paper]{article}
\usepackage{lrec2022} 
\usepackage{multibib}
\newcites{languageresource}{Language Resources}
\usepackage{graphicx}
\usepackage{tabularx}
\usepackage{soul}
\usepackage{titlesec}
\titleformat{\section}{\normalfont\large\bfseries\center}{\thesection.}{1em}{}
\titleformat{\subsection}{\normalfont\SmallTitleFont\bfseries\raggedright}{\thesubsection.}{1em}{}
\titleformat{\subsubsection}{\normalfont\normalsize\bfseries\raggedright}{\thesubsubsection.}{1em}{}
\renewcommand\thesection{\arabic{section}}
\renewcommand\thesubsection{\thesection.\arabic{subsection}}
\renewcommand\thesubsubsection{\thesubsection.\arabic{subsubsection}}

\usepackage{epstopdf}
\usepackage[utf8]{inputenc}

\usepackage{hyperref}
\usepackage{xstring}

\usepackage{color}

\usepackage[ethiop,main=english]{babel}
\usepackage{tipa}

\title{Extended Parallel Corpus for Amharic-English Machine Translation}

\name{Andargachew Mekonnen Gezmu, Andreas Nürnberger, Tesfaye Bayu Bati} 

\address{Otto von Guericke Universität Magdeburg \\
Universitätsplatz 2, Magdeburg, Germany \\
\{andargachew.gezmu, andreas.nuernberger\}@ovgu.de\\
tesfayebayu@hu.edu.et \\}

\abstract{
This paper describes the acquisition, preprocessing, segmentation, and alignment of an Amharic-English parallel corpus. It will be helpful for machine translation of a low-resource language, Amharic. We freely released the corpus for research purposes. Furthermore, we developed baseline statistical and neural machine translation systems; we trained statistical and neural machine translation models using the corpus. In the experiments, we also used a large monolingual corpus for the language model of statistical machine translation and back-translation of neural machine translation. In the automatic evaluation, neural machine translation models outperform statistical machine translation models by approximately six to seven Bilingual Evaluation Understudy (BLEU) points. Besides, among the neural machine translation models, the subword models outperform the word-based models by three to four BLEU points. Moreover, two other relevant automatic evaluation metrics, Translation Edit Rate on Character Level and Better Evaluation as Ranking, reflect corresponding differences among the trained models.
 \\ \newline  }

\begin{document}

\maketitleabstract

\section{Introduction}

To automate the intricate task of translation, researchers have followed different approaches. The earliest attempt was to use rule-based systems, which are criticized for being tedious and expensive to develop. Alternative empirical approaches such as Statistical Machine Translation (SMT) and Neural Machine Translation (NMT) came when parallel corpora were more and more available. Such methods take advantage of the authentic translations made by human translators in parallel corpora. They rely on machine learning to build translation models by taking parallel corpora as training data.

For empirical machine translation, we need a parallel corpus (bitext), a text that has a parallel translation in another language. Machine translation models are trained on a parallel corpus. Some international and governmental institutions provide such texts for public use. For instance, the Canadian Hansard corpus \citelanguageresource{roukoshansard} consists of parallel texts in English and French, drawn from official records of the proceedings of the Canadian Parliament. Similarly, the Europarl corpus \cite{DBLP:conf/acl/KoehnM05}, extracted from the proceedings of the European Parliament, contains parallel corpora for twenty-one European languages. The United Nations (UN) Parallel Corpus \cite{DBLP:conf/lrec/ZiemskiJP16} is available in six official UN languages. The current version of the parallel corpus consists of manually translated UN documents between 1990 and 2014.

Other parallel corpora have been made from movie subtitles, like the OpenSubtitles corpus \cite{DBLP:conf/lrec/LisonT16}, or from general web text, like the ParaCrawl corpus \cite{DBLP:conf/acl/BanonCHHHEFKKKO20}. The Open Parallel Corpus (OPUS) \cite{DBLP:conf/lrec/Tiedemann12} collects parallel corpora from sources such as open-source software documentations and religious books.

Large numbers of parallel corpora are available for dominant international languages such as English, German, and French. Nevertheless, there is a scarcity of available parallel corpora for low-resource languages. The deficiency impedes the progress of machine translation for such languages.

We considered different sources to develop a parallel corpus for Amharic-English translation. The existing corpora were either small or had poor quality; they were mainly collected from the web. Although considering the web as a corpus, which is motivated for practical reasons of getting more extensive data with open access and low cost, may sound good, such sources are inaccurate. Moreover, as Amharic is not standardized, one may face many spelling variations in these sources and expect typographical errors. This calls for manual or automatic editing. Therefore, we collected our corpora from edited documents such as newspapers, magazines, and textbooks. We also normalized the text and made some automatic spelling error corrections.

Amharic is a Semitic language that serves as the official language of Ethiopia. Although it plays several roles in the government, it is considered a low-resource language because of its lack of essential tools and resources for natural language processing \cite{DBLP:conf/lrec/GezmuNS18,DBLP:conf/sltu/TraceyS20}.

Amharic uses a syllabic writing system, Ethiopic. Each Amharic letter systematically conflates a consonant and vowel (\textit{e.g.}, \foreignlanguage{ethiop}{ba} /{b\textipa{@}}/ and \foreignlanguage{ethiop}{bu} /{bu}/). Sometimes consonants and vowels can be written as bare consonants (\textit{e.g.}, \foreignlanguage{ethiop}{be} /{b}/) or bare vowels (\textit{e.g.}, \foreignlanguage{ethiop}{'a} /{a}/ in \foreignlanguage{ethiop}{'agare} /{ag\textipa{@}r}/). Some phonemes with one or more homophonic script representations and peculiar labiovelars sometimes compromise the consistency of the writing system. In Amharic orthography, there is no case difference. It is written from left to right. In present-day Amharic writings, words are delimited by plain space.

Like other Semitic languages, Amharic words are highly inflectional and have a root-pattern morphology \cite{DBLP:series/tanlp/FabriGHKW14}. Therefore, Amharic lexicons cannot contain all word forms; the available bilingual lexicons contain only lemmas of common words. This hinders us to use sentence aligners that require bilingual lexicons.

Thus, in this research, we compiled a parallel corpus\footnote{The corpus is available at: \url{http://dx.doi.org/10.24352/ub.ovgu-2018-145}} for Amharic-English machine translation by extending the Ge'ez Frontier Foundation's news corpus made available for research purposes. We collected additional bilingual documents from various sources to compile the corpus. We also trained and evaluated NMT and SMT models using the corpus.

\section{Related Work}

There were attempts to compile parallel corpora for Amharic-English machine translation. The most notable ones are the ``Amharic-English bilingual corpus'', ``English-Ethiopian languages parallel corpora'' \cite{DBLP:conf/coling/AbateMTMAMAASAT18}, the ``Low Resource Languages for Emergent Incidents: Amharic representative language pack'' (LORELEI-Amharic) \cite{DBLP:conf/sltu/TraceyS20}, and the OPUS collection \cite{DBLP:conf/lrec/Tiedemann12}.

The European Language Resource Association (ELRA) hosts the Amharic-English bilingual corpus, containing a small parallel text from legal and news domains. \newcite{DBLP:conf/coling/AbateMTMAMAASAT18} compiled small-sized English-Ethiopian languages parallel corpora. Linguistic Data Consortium developed the LORELEI-Amharic corpus. Although LORELEI-Amharic is larger than Amharic-English bilingual corpus and English-Ethiopian languages parallel corpora, it is still not sufficient to train machine translation models with competitive performance \cite{DBLP:conf/aclnmt/KoehnK17,DBLP:conf/emnlp/LampleOCDR18}. Besides, the parallel text was collected from discussion forums, newswires, and weblogs. Discussion forums and weblogs are susceptible to spelling mistakes. The problem worsens as there is no readily available spell checker to assist Amharic writers.

In the OPUS collection, there are parallel corpora for Amharic and English. For this language pair, however, some of the corpora have a few hundred parallel sentences (\textit{e.g.}, Tatoeba, GlobalVoices, and TED2020); some use archaic language (\textit{e.g.}, Tanzil and bible-uedin); and others contain misaligned parallel sentences (\textit{e.g.}, MultiCCAligned and JW300).

The lack of clean, sizable, readily available, and contemporary-language parallel corpora impede the progress in Amharic-English machine translation. There were few attempts in Amharic-English machine translation using small-sized corpora \cite{DBLP:conf/sltu/TeshomeB12,teshome2015phoneme,DBLP:journals/mt/AshengoAA21}. Still, their corpora are not readily available for the research community.

\section{Corpus Preparation}
\label{sec:corpus}

We created a new parallel corpus by extending the existing news corpus made available for research purposes by Ge'ez Frontier Foundation. We collected, preprocessed, and segmented and aligned sentences of additional bilingual documents from various sources to compile the corpus.

\subsection{Data Sources}

We identified potential data sources that could serve as a basis for building a parallel corpus. We have considered newswires, magazines, and the Bible to get extensive data with open access.

Major newswires such as Deutsche Welle, BBC, and Ethiopian News Agency provide news articles in Amharic and English. Besides, the Ethiopian Herald and the Ethiopian Reporter publish bilingual news articles in Amharic and English. In these newswires, the translations are intended for the local public. Because of this, only a tiny portion of the English news articles are translated into Amharic, or vice versa. For instance, in the Ethiopian News Agency, approximately one news story out of ten has a rough translation \cite{DBLP:conf/webist/ArgawA05}.

Watchtower (\foreignlanguage{ethiop}{ma.tabaqiyA genebe} in Amharic) and Awake (\foreignlanguage{ethiop}{nequ} magazines in Amharic) have been published since 2006. They are available for the public; they have adequate sentence-by-sentence translations. Watchtower mainly discusses religious issues. Unlike Watchtower, Awake contains articles on general interest topics such as nature, geography, and family life. So it corresponds more to news articles.

The Bible is the most translated and readily available book. It is translated with great care and has high coverage of vocabulary \cite{chew-etal-2006-evaluation}. Additionally, its content reflects the everyday living of human beings like love, war, and politics. However, older translations of the Bible used archaic languages. Fortunately, we found out the recent translations of the Bible use the contemporary language. For example, the Standard Version and the New World Translation use the modern-day language in both Amharic and English.

Therefore, we selected text from Awake and Watchtower magazines, the Bible, and newswires. Then, we preprocessed the text as a preparation step for the following sentence segmentation and alignment activities.

\subsection{Preprocessing}

The preprocessing of the text involves spelling correction and normalization. In addition, we removed boilerplates such as headers, footers (including footnotes), and verse numbers (in the Bible).

In the text, we observed different types of misspellings: misspellings result from missed out spaces, replacing letters with visually similar characters (\textit{e.g.}, \foreignlanguage{ethiop}{qu} and \foreignlanguage{ethiop}{qui}), and typographical errors. We could not use our rule-based Amharic spelling corrector \cite{DBLP:conf/medes/Mekonnen12} because of its limitations. Instead, we developed another spelling corrector \cite{gezmu-etal-2018-contemporary} that has a better performance measured with the benchmark test set\footnote{The test set is available at: \url{https://github.com/andmek/ErrorCorpus}} \cite{DBLP:journals/corr/gezmu21err}. We employed the spelling corrector primarily to correct the first two types of spelling errors. Since the intensive manual intervention is needed to select the correct spelling from the plausible suggestions for typographical errors, we have not corrected the typographical errors in the current version of the corpus.

Different styles of punctuation marks have been used in Amharic text. For instance, for double quotation mark, two successive single quotation marks or similar symbols (\textit{e.g.}, \foreignlanguage{ethiop}{<}\foreignlanguage{ethiop}{<}, \foreignlanguage{ethiop}{>}\foreignlanguage{ethiop}{>}, \foreignlanguage{ethiop}{<<}, or \foreignlanguage{ethiop}{>>}) are used; for end-of-sentence punctuation (\foreignlanguage{ethiop}{::} ``Amharic full stop'') two successive Amharic word separator (\foreignlanguage{ethiop}{:}) that give the same appearance are used. Thus, the normalization of punctuation is a nontrivial matter. We normalized all types of double quotes, all single quotes, question marks (\textit{e.g.}, ? and \foreignlanguage{ethiop}{|}), word separators (\textit{e.g.}, : and \foreignlanguage{ethiop}{:}), full stops (\textit{e.g.}, :: and \foreignlanguage{ethiop}{::}), exclamation marks (\textit{e.g.}, \foreignlanguage{ethiop}{!} and !), hyphens (\textit{e.g.}, :- and \foreignlanguage{ethiop}{:-}), and commas (\textit{e.g.}, \foreignlanguage{ethiop}{:=} and \foreignlanguage{ethiop}{,}).

\subsection{Sentence Segmentation}

Segmentation of sentences essentially involves the disambiguation of end-of-sentence punctuation. To do so, we identified end-of-sentence punctuation marks. We considered end-of-sentence punctuation (\foreignlanguage{ethiop}{::} for Amharic and period for English) and question marks as a sentence boundary. The exceptions are abbreviations, initials of names, clitics, Uniform Resource Locators (URLs), e-mail addresses, and hashtags. Thus, to retain them we created a list of known abbreviations and clitics; and regular expressions for URLs, e-mail addresses, and hashtags. After sentence segmentation, we deleted duplicate sentences.

\subsection{Sentence Alignment}

Amharic has a rich morphology; it is practically impossible for Amharic lexicons to contain all word forms. Therefore, it is beneficial to use a sentence aligner that does not require any bilingual lexicon. Hence, we used the Bilingual Sentence Aligner\footnote{The implementation is available at: \url{https://www.microsoft.com/en-us/download/details.aspx?id=52608}} \cite{DBLP:conf/amta/Moore02} to align sentences in the bilingual documents. 

\begin{table}[!h]
\begin{center}
\begin{tabular}{|l|r|}
    \hline
    Document & Number of sentence pairs \\
    \hline
    Awake & 16491 \\
    \hline
    Watchtower & 72512 \\
    \hline
    The Bible & 48651 \\
    \hline
    News articles & 7710 \\
    \hline
    Total & 145364 \\
    \hline
\end{tabular}
\caption{The number of sentences (segments) aligned in each bilingual document.}
\label{tab:num_aligned_sents}
\end{center}
\end{table}

Table~\ref{tab:num_aligned_sents} shows the number of sentences aligned in each bilingual document. The corpus is comprised of approximately 83\% of the Watchtower magazine and Bible text that can be considered as a ``belief and thought'' domain \cite{burnard2007bnc}. The remaining 17\% of the Awake magazine and news articles is in the ``world affairs'' domain \cite{burnard2007bnc}.

\begin{table*}[!h]
\begin{center}
\begin{tabular}{|l|r|r|r|r|r|r|}
    \hline
    Dataset & Sentences & English Tokens & Amharic Tokens & English Types & Amharic Types \\
    \hline
    Test & 2500 & 46154 & 34689 & 5842 & 11644 \\
    \hline
    Validation & 2864 & 53818 & 39980 & 6470 & 13068 \\
    \hline
    Training & 140000 & 2574538 & 1930220 & 33589 & 155824 \\
    \hline
    Total & 145364 & 2674510 & 2004889 & 45901 & 180536 \\ 
    \hline
\end{tabular}
\caption{The number of sentences (segments), tokens, and types in each dataset.}
\label{tab:stat_dataset}
\end{center}
\end{table*}

After merging and shuffling the aligned sentences, we divided them into the training, validation (development), and test sets. Table~\ref{tab:stat_dataset} shows the statistics of each dataset.

\section{Baseline Systems}

During recent years, there are many improvements over SMT, such as hierarchical phrase-based SMT \cite{DBLP:journals/coling/Chiang07} and syntax-based SMT \cite{DBLP:conf/naacl/GalleyHKM04,DBLP:conf/acl/GalleyGKMDWT06}, and NMT like Universal Transformers \cite{DBLP:conf/iclr/DehghaniGVUK19}. Nevertheless, we relied on baseline systems for both approaches to evaluate them objectively.

\subsection{Baseline SMT System}

Our phrase-based SMT baseline system had settings that were typically used by \newcite{DBLP:conf/wmt/DingDKKP16}, \newcite{DBLP:conf/wmt/WilliamsSNHHB16}, \newcite{DBLP:conf/aclnmt/KoehnK17}, and \newcite{DBLP:conf/acl/SennrichZ19}. We used the Moses \cite{DBLP:conf/acl/KoehnHBCFBCSMZDBCH07} toolkit to train phrase-based SMT models. First, we used GIZA++ \cite{DBLP:conf/acl/Och03} and the grow-diag-final-and heuristic for symmetrization for word alignment. Then, we used the phrase-based reordering model \cite{DBLP:conf/naacl/KoehnOM03} with three different orientations: monotone, swap, and discontinuous in backward and forward directions conditioned on the source and target languages.

We used five-gram language models smoothed with the modified Kneser-Ney \cite{DBLP:conf/icassp/KneserN95}. The system applied KenLM \cite{DBLP:conf/wmt/Heafield11} language modeling toolkit for this purpose. Initially, we have not used big monolingual corpora for language models. This is because they are no longer the exclusive advantages of phrase-based SMT, as NMT can also benefit from them \cite{DBLP:conf/acl/SennrichZ19}. Afterward, to prove this claim, we used the Contemporary Amharic Corpus\footnote{The corpus is available at: \url{http://dx.doi.org/10.24352/ub.ovgu-2018-144}} (CACO) \cite{gezmu-etal-2018-contemporary} for English-to-Amharic translation.

The feature weights were tuned using Minimum Error Rate Training (MERT) \cite{DBLP:conf/acl/Och03}. We also used the k-best batch Margin Infused Relaxed Algorithm (MIRA) for tuning \cite{DBLP:conf/naacl/CherryF12} by selecting the highest-scoring development run with a return-best-dev setting.

In decoding, we applied the default normal stack search algorithm.

\subsection{Baseline NMT System}

To train NMT models, we used the encoder-decoder architecture implemented with Transformers \newcite{DBLP:conf/nips/VaswaniSPUJGKP17}. We tuned the hyperparameters of our NMT baseline system following \newcite{DBLP:conf/nips/VaswaniSPUJGKP17}, \newcite{DBLP:conf/wmt/DengCLSWWYZZZZC18}, \newcite{DBLP:conf/icaart/GezmuNB21}, and \newcite{DBLP:conf/icaart/GezmuN22}. The hyperparameters include the Adam optimizer \cite{DBLP:journals/corr/KingmaB14} with varied learning rate over the course of training, dropout \cite{DBLP:journals/jmlr/SrivastavaHKSS14} rate of 0.1, label smoothing \cite{DBLP:conf/cvpr/SzegedyVISW16} of value 0.1, batch size of 1024, six Transformer blocks with eight heads, filter size of 2048, and hidden size of 512. We used tensor2tensor \cite{DBLP:conf/amta/VaswaniBBCGGJKK18} library to implement the system.

The situation of training NMT models is complex because the training of NMT models is usually non-deterministic and hardly ever converges \cite{DBLP:journals/pbml/PopelB18}. Most research in NMT does not specify any stopping criteria. Some mention only an approximate number of days elapsed to train the models \cite{DBLP:journals/corr/BahdanauCB14} or the exact number of training steps \cite{DBLP:conf/nips/VaswaniSPUJGKP17}. We trained, thus, each NMT model for 250000 steps following \newcite{DBLP:conf/nips/VaswaniSPUJGKP17}. 

For decoding, we used a single model obtained by averaging the last twelve checkpoints. Following \newcite{DBLP:journals/corr/WuSCLNMKCGMKSJL16}, we used a beam search with a beam size of four and a length penalty of 0.6.

\section{Experiments and Evaluation}

We evaluated the performance of the SMT and NMT systems. We also made a comparison of word-based and subword-based NMT models. The experiments used the same datasets for each system.

\subsection{Datasets and Preprocessing}

We trained our models on the benchmark dataset -- the Amharic-English parallel corpus explained in Section~\ref{sec:corpus}. The training set consists of 140000 sentence pairs; the validation and test sets have 2864 and 2500 sentence pairs. 

We tokenized the English datasets with Moses' tokenizer script; we modified Moses' script to tokenize the Amharic datasets. Next, to share named-entities between the languages, the Amharic datasets were transliterated with a transliteration scheme, Amharic transliteration for machine translation\footnote{The implementation is available at: \url{https://github.com/andmek/AT4MT}}, which is fully discussed in \cite{DBLP:conf/icaart/GezmuNB21}.

We removed sentence pairs with extreme length ratios of more than one to nine and sentences longer than eighty tokens for the phrase-based SMT baseline. For word-based NMT models, we used a shared vocabulary of the top forty-four thousand most frequent tokens (tokens that appear five or more times in the corpus). We set this optimum vocabulary size because it will be too large to fit our GPU's memory if we include less frequent tokens. Besides, we used the word-piece method \cite{DBLP:conf/icassp/SchusterN12,DBLP:journals/corr/WuSCLNMKCGMKSJL16}, which is similar to Byte Pair Encoding \cite{gage1994new,DBLP:conf/acl/SennrichHB16a}, to segment words in the datasets for subword-based NMT models. We used the word-piece implementation in tensor2tensor\footnote{Available at: \url{https://github.com/tensorflow/tensor2tensor}} library. Furthermore, since the vocabulary size in word-piece has an impact on the performance of the NMT models, we trained models with different vocabulary sizes \cite{DBLP:journals/corr/WuSCLNMKCGMKSJL16,DBLP:conf/aclnmt/DenkowskiN17,DBLP:conf/emnlp/CherryFBFM18,DBLP:conf/wmt/DingDKKP16}.

\subsection{Evaluation}

Eventually, translation outputs of the test sets were detokenized and evaluated with a case-sensitive Bilingual Evaluation Understudy (BLEU) metric \cite{DBLP:conf/acl/PapineniRWZ02}. For consistency, we used the metric's implementation made by \newcite{DBLP:conf/wmt/Post18}, sacreBLEU\footnote{Signature BLEU+case.mixed+numrefs.1+smooth.exp+ tok.13a+version.1.4.9}. To fill the limitations of BLEU \cite{DBLP:conf/eacl/Callison-BurchOK06,DBLP:journals/coling/Reiter18}, we also used Better Evaluation as Ranking (BEER) \cite{DBLP:conf/wmt/StanojevicS14} and Translation Edit Rate on Character Level (CharacTER) \cite{DBLP:conf/wmt/WangPRN16} metrics. Unlike BLEU and BEER, the smaller the CharacTER score, the better. Moreover, the Amharic outputs were not back transliterated to use these automatic metrics effectively.

\begin{table*}[!h]
\begin{center}
\begin{tabular}{|l|l|r|r|r|}
  \hline 
  Translation Direction & System & BLEU & BEER & CharacTER \\
  \hline 
  Amharic-to-English & NMT-1K & 32.2 & 0.575 & 0.536 \\
    & NMT-2K & 32.2 & 0.575 & 0.536 \\
    & NMT-4K & 32.8 & 0.577 & 0.530 \\
    & NMT-8K & 33.0 & 0.576 & 0.527 \\
    & NMT-16K & 32.9 & 0.574 & 0.528 \\
    & NMT-32K & 32.2 & 0.570 & 0.539 \\
    & NMT-Word-Based & 28.8 & 0.537 & 0.588 \\
    & SMT-MERT & 26.0 & 0.514 & 0.629 \\
    & SMT-MIRA & 23.2 & 0.494 & 0.705 \\
  \hline 
  English-to-Amharic & NMT-1K & 25.5 & 0.558 & 0.520 \\
    & NMT-2K & 25.7 & 0.554 & 0.525 \\
    & NMT-4K & 26.1 & 0.557 & 0.517 \\
    & NMT-8K & 26.4 & 0.555 & 0.521 \\
    & NMT-16K & 26.7 & 0.555 & 0.520 \\
    & NMT-32K & 26.7 & 0.552 & 0.523 \\
    & NMT-Word-Based & 23.0 & 0.514 & 0.585 \\
    & SMT-MERT & 20.0 & 0.502 & 0.643 \\
    & SMT-MIRA & 19.2 & 0.484 & 0.704 \\
    \hline
\end{tabular}
\caption{Performance results of SMT and NMT models.}
\label{tab:arc_results}
\end{center}
\end{table*}

\section{Results}

Table~\ref{tab:arc_results} shows the performance results of the SMT and NMT systems with BLEU, BEER, and CharacTER metrics. The SMT system achieved better scores when feature weights were tuned using MERT than batch MIRA. Thus, we took the phrase-based SMT system tuned with MERT as our baseline. Likewise, the NMT baseline systems use vocabulary sizes of eight thousand (8K) and sixteen thousand (16K) in Amharic-to-English and English-to-Amharic translation directions. Example Amharic-to-English translation outputs are given at the appendix.

\begin{table}[!h]
\begin{center}
\begin{tabular}{|l|r|r|r|}
  \hline
  System & BLEU & BEER & CharacTER \\
  \hline
  SMT & 20.0 & 0.502 & 0.643 \\
  \hline
  SMT + CACO & 21.2 & 0.508 & 0.628 \\
  \hline
  NMT & 26.7 & 0.555 & 0.520 \\
  \hline
  NMT + CACO & 27.8 & 0.563 & 0.501 \\
  \hline
\end{tabular}
\caption{Performance results of English-to-Amharic translation using the CACO corpus.}
\label{tab:backtrans}
\end{center}
\end{table}

In both translation directions, NMT models with subword units score the highest values of all. The NMT baseline models outperform the SMT baseline models by
approximately six to seven BLEU. Among the NMT models, the subword-based models outperform the word-based models by three to four BLEU. The BEER and CharacTER metrics as well reflect corresponding differences.

Even though big monolingual corpora are not integral components of NMT, both SMT and NMT can benefit from them. Table~\ref{tab:backtrans} shows the results of English-to-Amharic translation using the CACO corpus for language model of the baseline SMT system, and back-translation of the baseline NMT system to produce synthetic training data \cite{DBLP:conf/acl/SennrichHB16,DBLP:conf/nips/HeXQWYLM16,DBLP:conf/acl/ChengXHHWSL16,DBLP:books/sp/Qin20}. Both systems gained more than one BLUE scores by using CACO. The baseline NMT model attained the optimum result when we randomly drew three times the size of the original training data from the CACO corpus and generated synthetic data by translating it to English.

\section{Conclusions}

We collected, preprocessed, segmented, and aligned Amharic-English parallel sentences from various sources. In doing so, we addressed different issues such as normalization and spelling correction. The corpus will be helpful for machine translation of a low-resource language, Amharic. Therefore, we freely released the corpus for research purposes. Also, we developed baseline SMT and NMT systems; we trained SMT and NMT models using the corpus. Additionally, we used a large monolingual corpus for the language model of SMT and back-translation of NMT in the experiments. As a result, NMT models outperform SMT models by approximately six to seven BLEU in the automatic evaluation. Besides, among the NMT models, the subword models outperform the word-based models by three to four BLEU. Moreover, two other relevant automatic evaluation metrics, CharacTER and BEER, reflect corresponding differences among the trained models.

We recommend future work to increase the size of the corpus by extracting text from scanned documents. In addition, we are engaged in doing additional experiments with other word segmentation methods for subword-based translations.

\section{Acknowledgements}

We want to thank Daniel Yacob, director of Ge'ez Frontier Foundation, for the Amharic news articles.

\section*{Appendix: Example Translation Outputs}

The following examples show the reference translations and the translated English sentences using NMT and SMT systems.\\

\textbf{Reference}: About that time, my parents asked me to come back home.\\
\textbf{NMT-Word-Based}: About that time, I asked my parents to go to their home.\\
\textbf{NMT-Subword-Based}: About that time, my parents asked me to return home.\\
\textbf{SMT-MERT}: About that time, my parents to return to the house to house work.\\
\textbf{SMT-MIRA}: About that time, my parents to return to the house.\\

\textbf{Reference}: 2. Can we really live forever?\\
\textbf{NMT-Word-Based}: 2. Can we really live forever?\\
\textbf{NMT-Subword-Based}: 2. Can we really live forever?\\
\textbf{SMT-MERT}: 2. Really, we can live forever?\\
\textbf{SMT-MIRA}: 2. Really live forever?\\

\textbf{Reference}: Sandra quickly discovered that she had been scammed.\\
\textbf{NMT-Word-Based}: Sandra saw that she was dying right away.\\
\textbf{NMT-Subword-Based}: Sandra immediately recognized that she was mistaken.\\
\textbf{SMT-MERT}: Sandra she immediately.\\
\textbf{SMT-MIRA}: Sandra immediately.\\

\textbf{Reference}: Distressing circumstances can have a terrible impact on us.\\
\textbf{NMT-Word-Based}: When distressing situations strike, they may feel emotionally.\\
\textbf{NMT-Subword-Based}: Distressing situations can cause anxiety.\\
\textbf{SMT-MERT}: When distressing situations can emotional.\\
\textbf{SMT-MIRA}: When distressing situations emotional.\\

\textbf{Reference}: Olive oil is used copiously, as it is produced there on a large scale.\\
\textbf{NMT-Word-Based}: Olive oil is used in abundant value.\\
\textbf{NMT-Subword-Based}: The olive oil is so extensive that it pushes on the abundant possible.\\
\textbf{SMT-MERT}: As the bulk of olive oil for the benefit of the.\\
\textbf{SMT-MIRA}: Olive oil as a in the.\\

\textbf{Reference}: Six years later, the whole world economy collapsed.\\
\textbf{NMT-Word-Based}: Six years later, the entire world economy was destroyed.\\
\textbf{NMT-Subword-Based}: Six years later, the global economy sank into the world.\\
\textbf{SMT-MERT}: Six years later, the entire world economy, have been shattered.\\
\textbf{SMT-MIRA}: Six years later the global economy, have been shattered.\\

\textbf{Reference}: We also need to remember that Jesus said: ``you must love your neighbor as yourself."\\
\textbf{NMT-Word-Based}: We must remember that Jesus state: ``you must love your neighbor as yourself."\\
\textbf{NMT-Subword-Based}: Keep in mind that Jesus also said: ``you must love your neighbor as yourself."\\
\textbf{SMT-MERT}: Jesus said: ``you must love your neighbor as yourself," that we keep in mind.\\
\textbf{SMT-MIRA}: Jesus said: ``you must love your neighbor as yourself" that we keep in mind.\\

\section{Bibliographical References}\label{reference}

\bibliographystyle{lrec2022-bib}
\bibliography{lrec2022-example}

\begin{thebibliography}{}

\bibitem[\protect\citename{{Roukos S. et al.}}1995]{roukoshansard}
{Roukos S. et al.}
\newblock (1995).
\newblock {\em {Hansard {F}rench/{E}nglish. {LDC} {C}atalog {N}o: LDC95T20}}.
\newblock Linguistic Data Consortium.

\end{thebibliography}


\begin{thebibliography}{}

\bibitem[\protect\citename{Abate \bgroup et al.\egroup
  }2018]{DBLP:conf/coling/AbateMTMAMAASAT18}
Abate, S.~T., Woldeyohannis, M.~M., Tachbelie, M.~Y., Meshesha, M., Atinafu,
  S., Mulugeta, W., Assabie, Y., Abera, H., Seyoum, B.~E., Abebe, T., Tsegaye,
  W., Lemma, A., Andargie, T., and Shifaw, S.
\newblock (2018).
\newblock Parallel corpora for bi-lingual english-ethiopian languages
  statistical machine translation.
\newblock In Emily~M. Bender, et~al., editors, {\em Proceedings of the 27th
  International Conference on Computational Linguistics, {COLING} 2018, Santa
  Fe, New Mexico, USA, August 20-26, 2018}, pages 3102--3111. Association for
  Computational Linguistics.

\bibitem[\protect\citename{Argaw and Asker}2005]{DBLP:conf/webist/ArgawA05}
Argaw, A.~A. and Asker, L.
\newblock (2005).
\newblock Web mining for an amharic - english bilingual corpus.
\newblock In Jos{\'{e}} Cordeiro, et~al., editors, {\em {WEBIST} 2005,
  Proceedings of the First International Conference on Web Information Systems
  and Technologies, Miami, USA, May 26-28, 2005}, pages 239--246. {INSTICC}
  Press.

\bibitem[\protect\citename{Ashengo \bgroup et al.\egroup
  }2021]{DBLP:journals/mt/AshengoAA21}
Ashengo, Y.~A., Aga, R.~T., and Abebe, S.~L.
\newblock (2021).
\newblock Context based machine translation with recurrent neural network for
  english-amharic translation.
\newblock {\em Mach. Transl.}, 35(1):19--36.

\bibitem[\protect\citename{Bahdanau \bgroup et al.\egroup
  }2015]{DBLP:journals/corr/BahdanauCB14}
Bahdanau, D., Cho, K., and Bengio, Y.
\newblock (2015).
\newblock Neural machine translation by jointly learning to align and
  translate.
\newblock In Yoshua Bengio et~al., editors, {\em 3rd International Conference
  on Learning Representations, {ICLR} 2015, San Diego, CA, USA, May 7-9, 2015,
  Conference Track Proceedings}.

\bibitem[\protect\citename{Ba{\~{n}}{\'{o}}n \bgroup et al.\egroup
  }2020]{DBLP:conf/acl/BanonCHHHEFKKKO20}
Ba{\~{n}}{\'{o}}n, M., Chen, P., Haddow, B., Heafield, K., Hoang, H.,
  Espl{\`{a}}{-}Gomis, M., Forcada, M.~L., Kamran, A., Kirefu, F., Koehn, P.,
  Ortiz{-}Rojas, S., Sempere, L.~P., Ram{\'{\i}}rez{-}S{\'{a}}nchez, G.,
  Sarr{\'{\i}}as, E., Strelec, M., Thompson, B., Waites, W., Wiggins, D., and
  Zaragoza, J.
\newblock (2020).
\newblock Paracrawl: Web-scale acquisition of parallel corpora.
\newblock In Dan Jurafsky, et~al., editors, {\em Proceedings of the 58th Annual
  Meeting of the Association for Computational Linguistics, {ACL} 2020, Online,
  July 5-10, 2020}, pages 4555--4567. Association for Computational
  Linguistics.

\bibitem[\protect\citename{Burnard}2007]{burnard2007bnc}
Burnard, L.
\newblock (2007).
\newblock Reference guide for the british national corpus (xml edition): Design
  of the corpus [online]. accessed 12 april 2022.

\bibitem[\protect\citename{Callison{-}Burch \bgroup et al.\egroup
  }2006]{DBLP:conf/eacl/Callison-BurchOK06}
Callison{-}Burch, C., Osborne, M., and Koehn, P.
\newblock (2006).
\newblock Re-evaluation the role of bleu in machine translation research.
\newblock In Diana McCarthy et~al., editors, {\em {EACL} 2006, 11st Conference
  of the European Chapter of the Association for Computational Linguistics,
  Proceedings of the Conference, April 3-7, 2006, Trento, Italy}. The
  Association for Computer Linguistics.

\bibitem[\protect\citename{Cheng \bgroup et al.\egroup
  }2016]{DBLP:conf/acl/ChengXHHWSL16}
Cheng, Y., Xu, W., He, Z., He, W., Wu, H., Sun, M., and Liu, Y.
\newblock (2016).
\newblock Semi-supervised learning for neural machine translation.
\newblock In {\em Proceedings of the 54th Annual Meeting of the Association for
  Computational Linguistics, {ACL} 2016, August 7-12, 2016, Berlin, Germany,
  Volume 1: Long Papers}. The Association for Computer Linguistics.

\bibitem[\protect\citename{Cherry and Foster}2012]{DBLP:conf/naacl/CherryF12}
Cherry, C. and Foster, G.~F.
\newblock (2012).
\newblock Batch tuning strategies for statistical machine translation.
\newblock In {\em Human Language Technologies: Conference of the North American
  Chapter of the Association of Computational Linguistics, Proceedings, June
  3-8, 2012, Montr{\'{e}}al, Canada}, pages 427--436. The Association for
  Computational Linguistics.

\bibitem[\protect\citename{Cherry \bgroup et al.\egroup
  }2018]{DBLP:conf/emnlp/CherryFBFM18}
Cherry, C., Foster, G.~F., Bapna, A., Firat, O., and Macherey, W.
\newblock (2018).
\newblock Revisiting character-based neural machine translation with capacity
  and compression.
\newblock In Ellen Riloff, et~al., editors, {\em Proceedings of the 2018
  Conference on Empirical Methods in Natural Language Processing, Brussels,
  Belgium, October 31 - November 4, 2018}, pages 4295--4305. Association for
  Computational Linguistics.

\bibitem[\protect\citename{Chew \bgroup et al.\egroup
  }2006]{chew-etal-2006-evaluation}
Chew, P.~A., Verzi, S.~J., Bauer, T.~L., and McClain, J.~T.
\newblock (2006).
\newblock Evaluation of the {B}ible as a resource for cross-language
  information retrieval.
\newblock In {\em Proceedings of the Workshop on Multilingual Language
  Resources and Interoperability}, pages 68--74, Sydney, Australia, July.
  Association for Computational Linguistics.

\bibitem[\protect\citename{Chiang}2007]{DBLP:journals/coling/Chiang07}
Chiang, D.
\newblock (2007).
\newblock Hierarchical phrase-based translation.
\newblock {\em Comput. Linguistics}, 33(2):201--228.

\bibitem[\protect\citename{Dehghani \bgroup et al.\egroup
  }2019]{DBLP:conf/iclr/DehghaniGVUK19}
Dehghani, M., Gouws, S., Vinyals, O., Uszkoreit, J., and Kaiser, L.
\newblock (2019).
\newblock Universal transformers.
\newblock In {\em 7th International Conference on Learning Representations,
  {ICLR} 2019, New Orleans, LA, USA, May 6-9, 2019}. OpenReview.net.

\bibitem[\protect\citename{Deng \bgroup et al.\egroup
  }2018]{DBLP:conf/wmt/DengCLSWWYZZZZC18}
Deng, Y., Cheng, S., Lu, J., Song, K., Wang, J., Wu, S., Yao, L., Zhang, G.,
  Zhang, H., Zhang, P., Zhu, C., and Chen, B.
\newblock (2018).
\newblock Alibaba's neural machine translation systems for {WMT18}.
\newblock In Ondrej Bojar, et~al., editors, {\em Proceedings of the Third
  Conference on Machine Translation: Shared Task Papers, {WMT} 2018, Belgium,
  Brussels, October 31 - November 1, 2018}, pages 368--376. Association for
  Computational Linguistics.

\bibitem[\protect\citename{Denkowski and
  Neubig}2017]{DBLP:conf/aclnmt/DenkowskiN17}
Denkowski, M.~J. and Neubig, G.
\newblock (2017).
\newblock Stronger baselines for trustable results in neural machine
  translation.
\newblock In Thang Luong, et~al., editors, {\em Proceedings of the First
  Workshop on Neural Machine Translation, NMT@ACL 2017, Vancouver, Canada,
  August 4, 2017}, pages 18--27. Association for Computational Linguistics.

\bibitem[\protect\citename{Ding \bgroup et al.\egroup
  }2016]{DBLP:conf/wmt/DingDKKP16}
Ding, S., Duh, K., Khayrallah, H., Koehn, P., and Post, M.
\newblock (2016).
\newblock The {JHU} machine translation systems for {WMT} 2016.
\newblock In {\em Proceedings of the First Conference on Machine Translation,
  {WMT} 2016, colocated with {ACL} 2016, August 11-12, Berlin, Germany}, pages
  272--280. The Association for Computer Linguistics.

\bibitem[\protect\citename{Fabri \bgroup et al.\egroup
  }2014]{DBLP:series/tanlp/FabriGHKW14}
Fabri, R., Gasser, M., Habash, N., Kiraz, G., and Wintner, S.
\newblock (2014).
\newblock Linguistic introduction: The orthography, morphology and syntax of
  semitic languages.
\newblock In Imed Zitouni, editor, {\em Natural Language Processing of Semitic
  Languages}, Theory and Applications of Natural Language Processing, pages
  3--41. Springer.

\bibitem[\protect\citename{Gage}1994]{gage1994new}
Gage, P.
\newblock (1994).
\newblock A new algorithm for data compression.
\newblock {\em C Users Journal}, 12(2):23--38.

\bibitem[\protect\citename{Galley \bgroup et al.\egroup
  }2004]{DBLP:conf/naacl/GalleyHKM04}
Galley, M., Hopkins, M., Knight, K., and Marcu, D.
\newblock (2004).
\newblock What's in a translation rule?
\newblock In Julia Hirschberg, et~al., editors, {\em Human Language Technology
  Conference of the North American Chapter of the Association for Computational
  Linguistics, {HLT-NAACL} 2004, Boston, Massachusetts, USA, May 2-7, 2004},
  pages 273--280. The Association for Computational Linguistics.

\bibitem[\protect\citename{Galley \bgroup et al.\egroup
  }2006]{DBLP:conf/acl/GalleyGKMDWT06}
Galley, M., Graehl, J., Knight, K., Marcu, D., DeNeefe, S., Wang, W., and
  Thayer, I.
\newblock (2006).
\newblock Scalable inference and training of context-rich syntactic translation
  models.
\newblock In Nicoletta Calzolari, et~al., editors, {\em {ACL} 2006, 21st
  International Conference on Computational Linguistics and 44th Annual Meeting
  of the Association for Computational Linguistics, Proceedings of the
  Conference, Sydney, Australia, 17-21 July 2006}. The Association for Computer
  Linguistics.

\bibitem[\protect\citename{Gezmu and
  N{\"{u}}rnberger}2022]{DBLP:conf/icaart/GezmuN22}
Gezmu, A.~M. and N{\"{u}}rnberger, A.
\newblock (2022).
\newblock Transformers for low-resource neural machine translation.
\newblock In Ana~Paula Rocha, et~al., editors, {\em Proceedings of the 14th
  International Conference on Agents and Artificial Intelligence, {ICAART}
  2022, Volume 1, Online Streaming, February 3-5, 2022}, pages 459--466.
  {SCITEPRESS}.

\bibitem[\protect\citename{Gezmu \bgroup et al.\egroup
  }2018a]{DBLP:conf/lrec/GezmuNS18}
Gezmu, A.~M., N{\"{u}}rnberger, A., and Seyoum, B.~E.
\newblock (2018a).
\newblock Portable spelling corrector for a less-resourced language: Amharic.
\newblock In Nicoletta Calzolari, et~al., editors, {\em Proceedings of the
  Eleventh International Conference on Language Resources and Evaluation,
  {LREC} 2018, Miyazaki, Japan, May 7-12, 2018}. European Language Resources
  Association {(ELRA)}.

\bibitem[\protect\citename{Gezmu \bgroup et al.\egroup
  }2018b]{gezmu-etal-2018-contemporary}
Gezmu, A.~M., Seyoum, B.~E., Gasser, M., and N{\"u}rnberger, A.
\newblock (2018b).
\newblock Contemporary {A}mharic corpus: Automatically morpho-syntactically
  tagged {A}mharic corpus.
\newblock In {\em Proceedings of the First Workshop on Linguistic Resources for
  Natural Language Processing}, pages 65--70, Santa Fe, New Mexico, USA,
  August. Association for Computational Linguistics.

\bibitem[\protect\citename{Gezmu \bgroup et al.\egroup
  }2021a]{DBLP:journals/corr/gezmu21err}
Gezmu, A.~M., Lema, T.~T., Seyoum, B.~E., and N{\"{u}}rnberger, A.
\newblock (2021a).
\newblock Manually annotated spelling error corpus for amharic.
\newblock {\em CoRR}, abs/2106.13521.

\bibitem[\protect\citename{Gezmu \bgroup et al.\egroup
  }2021b]{DBLP:conf/icaart/GezmuNB21}
Gezmu, A.~M., N{\"{u}}rnberger, A., and Bati, T.~B.
\newblock (2021b).
\newblock Neural machine translation for {A}mharic-{E}nglish translation.
\newblock In Ana~Paula Rocha, et~al., editors, {\em Proceedings of the 13th
  International Conference on Agents and Artificial Intelligence, {ICAART}
  2021, Volume 1, Online Streaming, February 4-6, 2021}, pages 526--532.
  {SCITEPRESS}.

\bibitem[\protect\citename{He \bgroup et al.\egroup
  }2016]{DBLP:conf/nips/HeXQWYLM16}
He, D., Xia, Y., Qin, T., Wang, L., Yu, N., Liu, T., and Ma, W.
\newblock (2016).
\newblock Dual learning for machine translation.
\newblock In Daniel~D. Lee, et~al., editors, {\em Advances in Neural
  Information Processing Systems 29: Annual Conference on Neural Information
  Processing Systems 2016, December 5-10, 2016, Barcelona, Spain}, pages
  820--828.

\bibitem[\protect\citename{Heafield}2011]{DBLP:conf/wmt/Heafield11}
Heafield, K.
\newblock (2011).
\newblock Kenlm: Faster and smaller language model queries.
\newblock In Chris Callison{-}Burch, et~al., editors, {\em Proceedings of the
  Sixth Workshop on Statistical Machine Translation, WMT@EMNLP 2011, Edinburgh,
  Scotland, UK, July 30-31, 2011}, pages 187--197. Association for
  Computational Linguistics.

\bibitem[\protect\citename{Kingma and Ba}2015]{DBLP:journals/corr/KingmaB14}
Kingma, D.~P. and Ba, J.
\newblock (2015).
\newblock Adam: {A} method for stochastic optimization.
\newblock In Yoshua Bengio et~al., editors, {\em 3rd International Conference
  on Learning Representations, {ICLR} 2015, San Diego, CA, USA, May 7-9, 2015,
  Conference Track Proceedings}.

\bibitem[\protect\citename{Kneser and Ney}1995]{DBLP:conf/icassp/KneserN95}
Kneser, R. and Ney, H.
\newblock (1995).
\newblock Improved backing-off for m-gram language modeling.
\newblock In {\em 1995 International Conference on Acoustics, Speech, and
  Signal Processing, {ICASSP} '95, Detroit, Michigan, USA, May 08-12, 1995},
  pages 181--184. {IEEE} Computer Society.

\bibitem[\protect\citename{Koehn and Knowles}2017]{DBLP:conf/aclnmt/KoehnK17}
Koehn, P. and Knowles, R.
\newblock (2017).
\newblock Six challenges for neural machine translation.
\newblock In Thang Luong, et~al., editors, {\em Proceedings of the First
  Workshop on Neural Machine Translation, NMT@ACL 2017, Vancouver, Canada,
  August 4, 2017}, pages 28--39. Association for Computational Linguistics.

\bibitem[\protect\citename{Koehn and Monz}2005]{DBLP:conf/acl/KoehnM05}
Koehn, P. and Monz, C.
\newblock (2005).
\newblock Shared task: Statistical machine translation between european
  languages.
\newblock In Philipp Koehn, et~al., editors, {\em Proceedings of the Workshop
  on Building and Using Parallel Texts@ACL 2005, Ann Arbor, Michigan, USA, June
  29-30, 2005}, pages 119--124. Association for Computational Linguistics.

\bibitem[\protect\citename{Koehn \bgroup et al.\egroup
  }2003]{DBLP:conf/naacl/KoehnOM03}
Koehn, P., Och, F.~J., and Marcu, D.
\newblock (2003).
\newblock Statistical phrase-based translation.
\newblock In Marti~A. Hearst et~al., editors, {\em Human Language Technology
  Conference of the North American Chapter of the Association for Computational
  Linguistics, {HLT-NAACL} 2003, Edmonton, Canada, May 27 - June 1, 2003}. The
  Association for Computational Linguistics.

\bibitem[\protect\citename{Koehn \bgroup et al.\egroup
  }2007]{DBLP:conf/acl/KoehnHBCFBCSMZDBCH07}
Koehn, P., Hoang, H., Birch, A., Callison{-}Burch, C., Federico, M., Bertoldi,
  N., Cowan, B., Shen, W., Moran, C., Zens, R., Dyer, C., Bojar, O.,
  Constantin, A., and Herbst, E.
\newblock (2007).
\newblock Moses: Open source toolkit for statistical machine translation.
\newblock In John~A. Carroll, et~al., editors, {\em {ACL} 2007, Proceedings of
  the 45th Annual Meeting of the Association for Computational Linguistics,
  June 23-30, 2007, Prague, Czech Republic}. The Association for Computational
  Linguistics.

\bibitem[\protect\citename{Lample \bgroup et al.\egroup
  }2018]{DBLP:conf/emnlp/LampleOCDR18}
Lample, G., Ott, M., Conneau, A., Denoyer, L., and Ranzato, M.
\newblock (2018).
\newblock Phrase-based {\&} neural unsupervised machine translation.
\newblock In Ellen Riloff, et~al., editors, {\em Proceedings of the 2018
  Conference on Empirical Methods in Natural Language Processing, Brussels,
  Belgium, October 31 - November 4, 2018}, pages 5039--5049. Association for
  Computational Linguistics.

\bibitem[\protect\citename{Lison and Tiedemann}2016]{DBLP:conf/lrec/LisonT16}
Lison, P. and Tiedemann, J.
\newblock (2016).
\newblock Opensubtitles2016: Extracting large parallel corpora from movie and
  {TV} subtitles.
\newblock In Nicoletta Calzolari, et~al., editors, {\em Proceedings of the
  Tenth International Conference on Language Resources and Evaluation {LREC}
  2016, Portoro{\v{z}}, Slovenia, May 23-28, 2016}. European Language Resources
  Association {(ELRA)}.

\bibitem[\protect\citename{Mekonnen}2012]{DBLP:conf/medes/Mekonnen12}
Mekonnen, A.
\newblock (2012).
\newblock Development of an amharic spelling corrector for tolerant-retrieval.
\newblock In Janusz Kacprzyk, et~al., editors, {\em International Conference on
  Management of Emergent Digital EcoSystems, {MEDES} '12, Addis Ababa,
  Ethiopia, October 28-31, 2012}, pages 22--26. {ACM}.

\bibitem[\protect\citename{Moore}2002]{DBLP:conf/amta/Moore02}
Moore, R.~C.
\newblock (2002).
\newblock Fast and accurate sentence alignment of bilingual corpora.
\newblock In Stephen~D. Richardson, editor, {\em Machine Translation: From
  Research to Real Users, 5th Conference of the Association for Machine
  Translation in the Americas, {AMTA} 2002 Tiburon, CA, USA, October 6-12,
  2002, Proceedings}, volume 2499 of {\em Lecture Notes in Computer Science},
  pages 135--144. Springer.

\bibitem[\protect\citename{Och}2003]{DBLP:conf/acl/Och03}
Och, F.~J.
\newblock (2003).
\newblock Minimum error rate training in statistical machine translation.
\newblock In Erhard~W. Hinrichs et~al., editors, {\em Proceedings of the 41st
  Annual Meeting of the Association for Computational Linguistics, 7-12 July
  2003, Sapporo Convention Center, Sapporo, Japan}, pages 160--167. {ACL}.

\bibitem[\protect\citename{Papineni \bgroup et al.\egroup
  }2002]{DBLP:conf/acl/PapineniRWZ02}
Papineni, K., Roukos, S., Ward, T., and Zhu, W.
\newblock (2002).
\newblock Bleu: a method for automatic evaluation of machine translation.
\newblock In {\em Proceedings of the 40th Annual Meeting of the Association for
  Computational Linguistics, July 6-12, 2002, Philadelphia, PA, {USA}}, pages
  311--318. {ACL}.

\bibitem[\protect\citename{Popel and Bojar}2018]{DBLP:journals/pbml/PopelB18}
Popel, M. and Bojar, O.
\newblock (2018).
\newblock Training tips for the transformer model.
\newblock {\em Prague Bull. Math. Linguistics}, 110:43--70.

\bibitem[\protect\citename{Post}2018]{DBLP:conf/wmt/Post18}
Post, M.
\newblock (2018).
\newblock A call for clarity in reporting {BLEU} scores.
\newblock In Ondrej Bojar, et~al., editors, {\em Proceedings of the Third
  Conference on Machine Translation: Research Papers, {WMT} 2018, Belgium,
  Brussels, October 31 - November 1, 2018}, pages 186--191. Association for
  Computational Linguistics.

\bibitem[\protect\citename{Qin}2020]{DBLP:books/sp/Qin20}
Qin, T.
\newblock (2020).
\newblock {\em Dual Learning}.
\newblock Springer.

\bibitem[\protect\citename{Reiter}2018]{DBLP:journals/coling/Reiter18}
Reiter, E.
\newblock (2018).
\newblock A structured review of the validity of {BLEU}.
\newblock {\em Comput. Linguistics}, 44(3).

\bibitem[\protect\citename{Schuster and
  Nakajima}2012]{DBLP:conf/icassp/SchusterN12}
Schuster, M. and Nakajima, K.
\newblock (2012).
\newblock Japanese and korean voice search.
\newblock In {\em 2012 {IEEE} International Conference on Acoustics, Speech and
  Signal Processing, {ICASSP} 2012, Kyoto, Japan, March 25-30, 2012}, pages
  5149--5152. {IEEE}.

\bibitem[\protect\citename{Sennrich and Zhang}2019]{DBLP:conf/acl/SennrichZ19}
Sennrich, R. and Zhang, B.
\newblock (2019).
\newblock Revisiting low-resource neural machine translation: {A} case study.
\newblock In Anna Korhonen, et~al., editors, {\em Proceedings of the 57th
  Conference of the Association for Computational Linguistics, {ACL} 2019,
  Florence, Italy, July 28- August 2, 2019, Volume 1: Long Papers}, pages
  211--221. Association for Computational Linguistics.

\bibitem[\protect\citename{Sennrich \bgroup et al.\egroup
  }2016a]{DBLP:conf/acl/SennrichHB16}
Sennrich, R., Haddow, B., and Birch, A.
\newblock (2016a).
\newblock Improving neural machine translation models with monolingual data.
\newblock In {\em Proceedings of the 54th Annual Meeting of the Association for
  Computational Linguistics, {ACL} 2016, August 7-12, 2016, Berlin, Germany,
  Volume 1: Long Papers}. The Association for Computer Linguistics.

\bibitem[\protect\citename{Sennrich \bgroup et al.\egroup
  }2016b]{DBLP:conf/acl/SennrichHB16a}
Sennrich, R., Haddow, B., and Birch, A.
\newblock (2016b).
\newblock Neural machine translation of rare words with subword units.
\newblock In {\em Proceedings of the 54th Annual Meeting of the Association for
  Computational Linguistics, {ACL} 2016, August 7-12, 2016, Berlin, Germany,
  Volume 1: Long Papers}. The Association for Computer Linguistics.

\bibitem[\protect\citename{Srivastava \bgroup et al.\egroup
  }2014]{DBLP:journals/jmlr/SrivastavaHKSS14}
Srivastava, N., Hinton, G.~E., Krizhevsky, A., Sutskever, I., and
  Salakhutdinov, R.
\newblock (2014).
\newblock Dropout: a simple way to prevent neural networks from overfitting.
\newblock {\em J. Mach. Learn. Res.}, 15(1):1929--1958.

\bibitem[\protect\citename{Stanojevic and
  Sima'an}2014]{DBLP:conf/wmt/StanojevicS14}
Stanojevic, M. and Sima'an, K.
\newblock (2014).
\newblock {BEER:} better evaluation as ranking.
\newblock In {\em Proceedings of the Ninth Workshop on Statistical Machine
  Translation, WMT@ACL 2014, June 26-27, 2014, Baltimore, Maryland, {USA}},
  pages 414--419. The Association for Computer Linguistics.

\bibitem[\protect\citename{Szegedy \bgroup et al.\egroup
  }2016]{DBLP:conf/cvpr/SzegedyVISW16}
Szegedy, C., Vanhoucke, V., Ioffe, S., Shlens, J., and Wojna, Z.
\newblock (2016).
\newblock Rethinking the inception architecture for computer vision.
\newblock In {\em 2016 {IEEE} Conference on Computer Vision and Pattern
  Recognition, {CVPR} 2016, Las Vegas, NV, USA, June 27-30, 2016}, pages
  2818--2826. {IEEE} Computer Society.

\bibitem[\protect\citename{Teshome and
  Besacier}2012]{DBLP:conf/sltu/TeshomeB12}
Teshome, M.~G. and Besacier, L.
\newblock (2012).
\newblock Preliminary experiments on english-amharic statistical machine
  translation.
\newblock In {\em Third Workshop on Spoken Language Technologies for
  Under-resourced Languages, {SLTU} 2012, Cape Town, South Africa, May 7-9,
  2012}, pages 36--41. {ISCA}.

\bibitem[\protect\citename{Teshome \bgroup et al.\egroup
  }2015]{teshome2015phoneme}
Teshome, M.~G., Besacier, L., Taye, G., and Teferi, D.
\newblock (2015).
\newblock {P}honeme-based {E}nglish-{A}mharic statistical machine translation.
\newblock In {\em AFRICON 2015}, pages 1--5. IEEE.

\bibitem[\protect\citename{Tiedemann}2012]{DBLP:conf/lrec/Tiedemann12}
Tiedemann, J.
\newblock (2012).
\newblock Parallel data, tools and interfaces in {OPUS}.
\newblock In Nicoletta Calzolari, et~al., editors, {\em Proceedings of the
  Eighth International Conference on Language Resources and Evaluation, {LREC}
  2012, Istanbul, Turkey, May 23-25, 2012}, pages 2214--2218. European Language
  Resources Association {(ELRA)}.

\bibitem[\protect\citename{Tracey and Strassel}2020]{DBLP:conf/sltu/TraceyS20}
Tracey, J. and Strassel, S.~M.
\newblock (2020).
\newblock Basic language resources for 31 languages (plus english): The
  {LORELEI} representative and incident language packs.
\newblock In Dorothee Beermann, et~al., editors, {\em Proceedings of the 1st
  Joint Workshop on Spoken Language Technologies for Under-resourced languages
  and Collaboration and Computing for Under-Resourced Languages,
  SLTU/CCURL@LREC 2020, Marseille, France, May 2020}, pages 277--284. European
  Language Resources association.

\bibitem[\protect\citename{Vaswani \bgroup et al.\egroup
  }2017]{DBLP:conf/nips/VaswaniSPUJGKP17}
Vaswani, A., Shazeer, N., Parmar, N., Uszkoreit, J., Jones, L., Gomez, A.~N.,
  Kaiser, L., and Polosukhin, I.
\newblock (2017).
\newblock Attention is all you need.
\newblock In Isabelle Guyon, et~al., editors, {\em Advances in Neural
  Information Processing Systems 30: Annual Conference on Neural Information
  Processing Systems 2017, December 4-9, 2017, Long Beach, CA, {USA}}, pages
  5998--6008.

\bibitem[\protect\citename{Vaswani \bgroup et al.\egroup
  }2018]{DBLP:conf/amta/VaswaniBBCGGJKK18}
Vaswani, A., Bengio, S., Brevdo, E., Chollet, F., Gomez, A.~N., Gouws, S.,
  Jones, L., Kaiser, L., Kalchbrenner, N., Parmar, N., Sepassi, R., Shazeer,
  N., and Uszkoreit, J.
\newblock (2018).
\newblock Tensor2tensor for neural machine translation.
\newblock In Colin Cherry et~al., editors, {\em Proceedings of the 13th
  Conference of the Association for Machine Translation in the Americas, {AMTA}
  2018, Boston, MA, USA, March 17-21, 2018 - Volume 1: Research Papers}, pages
  193--199. Association for Machine Translation in the Americas.

\bibitem[\protect\citename{Wang \bgroup et al.\egroup
  }2016]{DBLP:conf/wmt/WangPRN16}
Wang, W., Peter, J., Rosendahl, H., and Ney, H.
\newblock (2016).
\newblock Character: Translation edit rate on character level.
\newblock In {\em Proceedings of the First Conference on Machine Translation,
  {WMT} 2016, colocated with {ACL} 2016, August 11-12, Berlin, Germany}, pages
  505--510. The Association for Computer Linguistics.

\bibitem[\protect\citename{Williams \bgroup et al.\egroup
  }2016]{DBLP:conf/wmt/WilliamsSNHHB16}
Williams, P., Sennrich, R., Nadejde, M., Huck, M., Haddow, B., and Bojar, O.
\newblock (2016).
\newblock Edinburgh's statistical machine translation systems for {WMT16}.
\newblock In {\em Proceedings of the First Conference on Machine Translation,
  {WMT} 2016, colocated with {ACL} 2016, August 11-12, Berlin, Germany}, pages
  399--410. The Association for Computer Linguistics.

\bibitem[\protect\citename{Wu \bgroup et al.\egroup
  }2016]{DBLP:journals/corr/WuSCLNMKCGMKSJL16}
Wu, Y., Schuster, M., Chen, Z., Le, Q.~V., Norouzi, M., Macherey, W., Krikun,
  M., Cao, Y., Gao, Q., Macherey, K., Klingner, J., Shah, A., Johnson, M., Liu,
  X., Kaiser, L., Gouws, S., Kato, Y., Kudo, T., Kazawa, H., Stevens, K.,
  Kurian, G., Patil, N., Wang, W., Young, C., Smith, J., Riesa, J., Rudnick,
  A., Vinyals, O., Corrado, G., Hughes, M., and Dean, J.
\newblock (2016).
\newblock Google's neural machine translation system: Bridging the gap between
  human and machine translation.
\newblock {\em CoRR}, abs/1609.08144.

\bibitem[\protect\citename{Ziemski \bgroup et al.\egroup
  }2016]{DBLP:conf/lrec/ZiemskiJP16}
Ziemski, M., Junczys{-}Dowmunt, M., and Pouliquen, B.
\newblock (2016).
\newblock The united nations parallel corpus v1.0.
\newblock In Nicoletta Calzolari, et~al., editors, {\em Proceedings of the
  Tenth International Conference on Language Resources and Evaluation {LREC}
  2016, Portoro{\v{z}}, Slovenia, May 23-28, 2016}. European Language Resources
  Association {(ELRA)}.

\end{thebibliography}

\section{Language Resource References}
\label{lr:ref}
\bibliographystylelanguageresource{lrec2022-bib}
\bibliographylanguageresource{languageresource}

\end{document}